\documentclass[11pt,a4paper]{article}
\usepackage{times}
\usepackage{latexsym}
\PassOptionsToPackage{breaklinks}{hyperref}
\usepackage[hyperref]{acl2021}

\usepackage{microtype}

\usepackage{multirow}
\usepackage{booktabs}
\usepackage{graphicx}
\usepackage{subfigure}
\usepackage{amssymb}
\usepackage{CJKutf8}
\usepackage{amsmath} 
\usepackage{hyperref}
\usepackage{xurl}
\usepackage{enumitem}
\usepackage{color}

\newcommand{\pchn}[1]{\protect\begin{CJK*}{UTF8}{gbsn}#1\protect\end{CJK*}}

\aclfinalcopy 
\def\aclpaperid{2344} 

\title{Lexicon Enhanced Chinese Sequence Labelling Using BERT Adapter}

 \author{Wei Liu$^{1}$, Xiyan Fu$^{2}$, Yue Zhang$^{3}$, Wenming Xiao$^{1}$ \\ $^{1}$DAMO Academy, Alibaba Group, China  \\ $^{2}$College of Computer Science, Nankai University, China \\ $^{3}$School of Engineering, Westlake University, China \\ $^{3}$Institute of Advanced Technology, Westlake Institute for Advanced Study
 \\ \texttt{hezan.lw@alibaba-inc.com}, \texttt{fuxiyan@mail.nankai.edu.cn}, \\ \texttt{yue.zhang@wias.org.cn}, \texttt{wenming.xiaowm@alibaba-inc.com}}

\date{}

\begin{document}
\maketitle

\begin{abstract}
Lexicon information and pre-trained models, such as BERT, have been combined to explore Chinese sequence labeling tasks due to their respective strengths. However, existing methods solely fuse lexicon features via a shallow and random initialized sequence layer and do not integrate them into the bottom layers of BERT. In this paper, we propose Lexicon Enhanced BERT (LEBERT) for Chinese sequence labeling, which integrates external lexicon knowledge into BERT layers directly by a Lexicon Adapter layer. Compared with existing methods, our model facilitates deep lexicon knowledge fusion at the lower layers of BERT. Experiments on ten Chinese datasets of three tasks including Named Entity Recognition, Word Segmentation, and Part-of-Speech Tagging, show that LEBERT achieves state-of-the-art results.

\end{abstract}

\section{Introduction}
Sequence labeling is a classic task in natural language processing (NLP), which is to assign a label to each unit in a sequence \cite{speech-language}. Many important language processing tasks can be converted into this problem, such as part-of-speech (POS) tagging, named entity recognition (NER), and text chunking. The current state-of-the-art results for sequence labeling have been achieved by neural network approaches ~\citep{lample-etal-2016-neural, ma-hovy-2016-end, TACL792, gui-etal-2017-part}. 

Chinese sequence labeling is more challenging due to the lack of explicit word boundaries in Chinese sentences. One way of performing Chinese sequence labeling is to perform Chinese word segmentation (CWS) first, before applying word sequence labeling ~\citep{sun-uszkoreit-2012-capturing,yang2016combining}. However, it can suffer from the segmentation errors propagated from the CWS system ~\cite{zhang-yang-2018-chinese,liu-etal-2019-encoding}. Therefore, some approaches ~\citep{cao-etal-2018-adversarial,shen-etal-2016-consistent} perform Chinese sequence labeling directly at the character level, which has been empirically proven to be more effective ~\citep{ng-low-2004-chinese,2010Chinese,zhang-yang-2018-chinese}.

\begin{figure}
    \centering
    \subfigure[Model-level Fusion]{
        \centering
        \includegraphics[scale=0.31]{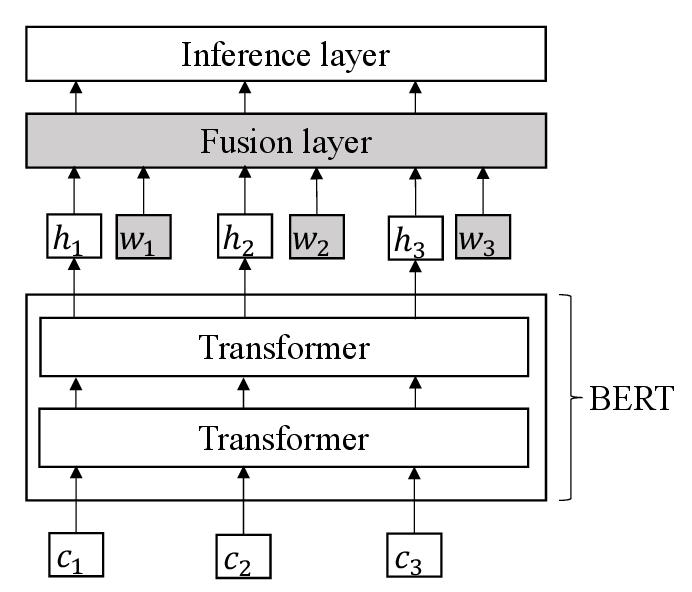}
        \label{fig-compare-a}
    }
    \subfigure[BERT-level Fusion]{
        \centering
        \includegraphics[scale=0.28]{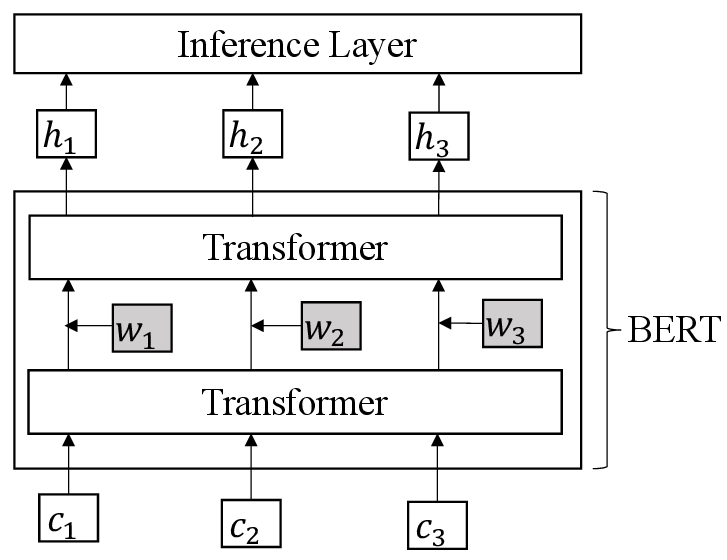}
        \label{fig-compare-b}
    }
    \caption{Comparison of fusing lexicon features and BERT at different levels for Chinese sequence labeling. For simplicity, we only show two Transformer layers in BERT and truncate the sentence to three characters. $c_i$ denotes the $i$-th Chinese character, $w_j$ denotes the $j$-th Chinese word.}
    \label{fig-compare}
\end{figure}

There are two lines of recent work enhancing character-based neural Chinese sequence labeling. The first considers integrating word information into a character-based sequence encoder, so that word features can be explicitly modeled ~\citep{zhang-yang-2018-chinese,yang-etal-2019-subword,liu-etal-2019-encoding,ding-etal-2019-neural,higashiyama-etal-2019-incorporating}. These methods can be treated as designing different variants to neural architectures for integrating discrete structured knowledge. The second considers the integration of large-scale pre-trained contextualized embeddings, such as BERT ~\citep{devlin-etal-2019-bert}, which has been shown to capture implicit word-level syntactic and semantic knowledge ~\citep{assessBERT,hewitt-manning-2019-structural}.

The two lines of work are complementary to each other due to the different nature of discrete and neural representations. Recent work considers the combination of lexicon features and BERT for Chinese NER ~\citep{ma-etal-2020-simplify,li-etal-2020-flat}, Chinese Word Segmentation \cite{ganlan}, and Chinese POS tagging ~\citep{tian-etal-2020-joint}. The main idea is to integrate contextual representations from BERT and lexicon features into a neural sequence labeling model (shown in Figure 1 \subref{fig-compare-a}). However, these approaches do not fully exploit the representation power of BERT, because the external features are not integrated into the bottom level.

Inspired by the work about BERT Adapter ~\citep{houlsby2019parameter, bapna-firat-2019-simple, wang2020kadapter}, we propose Lexicon Enhanced BERT (LEBERT) to integrate lexicon information between Transformer layers of BERT directly. Specifically, a Chinese sentence is converted into a char-words pair sequence by matching the sentence with an existing lexicon. A lexicon adapter is designed to dynamically extract the most relevant matched words for each character using a char-to-word bilinear attention mechanism. The lexicon adapter is applied between adjacent transformers in BERT (shown in Figure 1 \subref{fig-compare-b}) so that lexicon features and BERT representation interact sufficiently through the multi-layer encoder within BERT. We fine-tune both the BERT and lexicon adapter during training to make full use of word information, which is considerably different from the BERT Adapter (it fixes BERT parameters). 

We investigate the effectiveness of LEBERT on three Chinese sequence labeling tasks\footnote{\url{https://github.com/liuwei1206/LEBERT}}, including Chinese NER, Chinese Word Segmentation\footnote{We follow the mainstream methods and regard Chinese Word Segmentation as a sequence labeling problem.}, and Chinese POS tagging. Experimental results on ten benchmark datasets illustrate the effectiveness of our model, where state-of-the-art performance is achieved for each task on all datasets. In addition, we provide comprehensive comparisons and detailed analyses, which empirically confirm that bottom-level feature integration contributes to span boundary detection and span type determination.

\section{Related Work}
Our work is related to existing neural methods using lexicon features and pre-trained models to improve Chinese sequence labeling.

\noindent{\textbf{Lexicon-based}.} Lexicon-based models aim to enhance character-based models with lexicon information. \citet{zhang-yang-2018-chinese} introduced a lattice LSTM to encode both characters and words for Chinese NER. It is further improved by following efforts in terms of training efficiency~\citep{ijcai2019-692,ma-etal-2020-simplify}, model degradation \cite{liu-etal-2019-encoding}, graph structure~\citep{gui-etal-2019-lexicon,ding-etal-2019-neural}, and removing the dependency of the lexicon \cite{zhu-wang-2019-ner}. Lexicon information has also been shown helpful for Chinese Word Segmentation (CWS) and Part-of-speech (POS) tagging. \citet{yang-etal-2019-subword} applied a lattice LSTM for CWS, showing good performance. \citet{SIGIRzhao} improved the results of CWS with lexicon-enhanced adaptive attention. \citet{tian-etal-2020-joint} enhanced the character-based Chinese POS tagging model with a multi-channel attention of N-grams.

\noindent{\textbf{{Pre-trained Model-based}}.} Transformer-based pre-trained models, such as BERT ~\citep{devlin-etal-2019-bert}, have shown excellent performance for Chinese sequence labeling. \citet{BERTmeetCWS} simply added a softmax on BERT, achieving state-of-the-art performance on CWS.  \citet{NEURIPS2019_452bf208, hu-verberne-2020-named} showed that models using the character features from BERT outperform the static embedding-based approaches by a large margin for Chinese NER and Chinese POS tagging.

\noindent\textbf{Hybrid Model}. Recent work tries to integrate the lexicon and pre-trained models by utilizing their respective strengths. \citet{ma-etal-2020-simplify} concatenated separate features, BERT representation and lexicon information, and input them into a shallow fusion layer (LSTM) for Chinese NER. \citet{li-etal-2020-flat} proposed a shallow Flat-Lattice Transformer to handle the character-word graph, in which the fusion is still at model-level. Similarly, character N-gram features and BERT vectors are concatenated for joint training CWS and POS tagging ~\cite{tian-etal-2020-joint}. Our method is in line with the above approaches trying to combine lexicon information and BERT. The difference is that we integrate lexicon into the bottom level, allowing in-depth knowledge interaction within BERT. 

\begin{figure}[ht]
\centering\includegraphics[scale=0.29,trim=3 0 0 0]{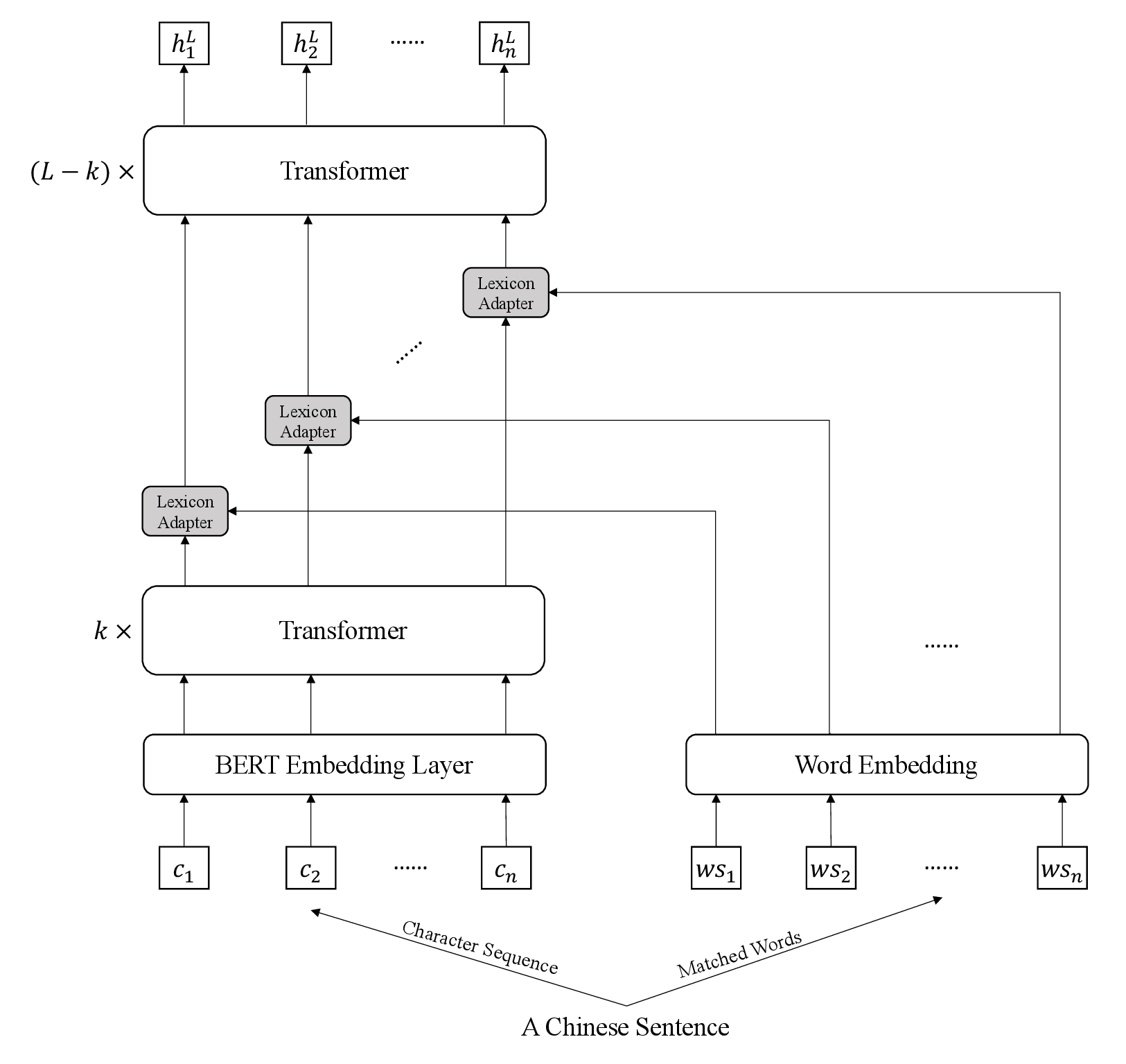}
\setlength{\abovecaptionskip}{1pt} 
\caption{The architecture of Lexicon Enhanced BERT, in which lexicon features are integrated between $k$-th and $(k+1)$-th Transformer Layer using Lexicon Adapter. Where $c_i$ denote the $i$-th Chinese character in the sentence, and ${ws}_i$ denotes matched words assigned to character $c_i$.}
\label{fig_model}
\vspace{-0.1in}
\end{figure}

There is also work employing lexicon to guide pre-training. ERNIE ~\citep{sun2019ernie, sun2019ernie20} exploited entity-level and word-level masking to integrate knowledge into BERT in an implicit way. \newcite{jia-etal-2020-entity} proposed Entity Enhanced BERT, further pre-training BERT using a domain-specific corpus and entity set with a carefully designed character-entity Transformer. ZEN \cite{diao-etal-2020-zen} enhanced Chinese BERT with a multi-layered N-gram encoder but is limited by the small size of the N-gram vocabulary. Compared to the above pre-training methods, our model integrates lexicon information into BERT using an adapter, which is more efficient and requires no raw texts or entity set.

\noindent \textbf{BERT Adapter}. BERT Adapter~\cite{houlsby2019parameter} aims to learn task-specific parameters for the downstream tasks. Specifically, they add adapters between layers of a pre-trained model and tune only the parameters in the added adapters for a certain task. \citet{bapna-firat-2019-simple} injected task-specific adapter layers into pre-trained models for neural machine translation. MAD-X \cite{pfeiffer2020madx} is an adapter-based framework that enables high portability and parameter-efficient transfer to arbitrary tasks. \citet{wang2020kadapter} proposed K-ADAPTER to infuse knowledge into pre-trained models with further pre-training. Similar to them, we use a lexicon adapter to integrate lexicon information into BERT. The main difference is that our goal is to better fuse lexicon and BERT at the bottom-level rather than efficient training. 
\begin{figure}[ht]
\centering\includegraphics[scale=0.3,trim=0 0 0 0]{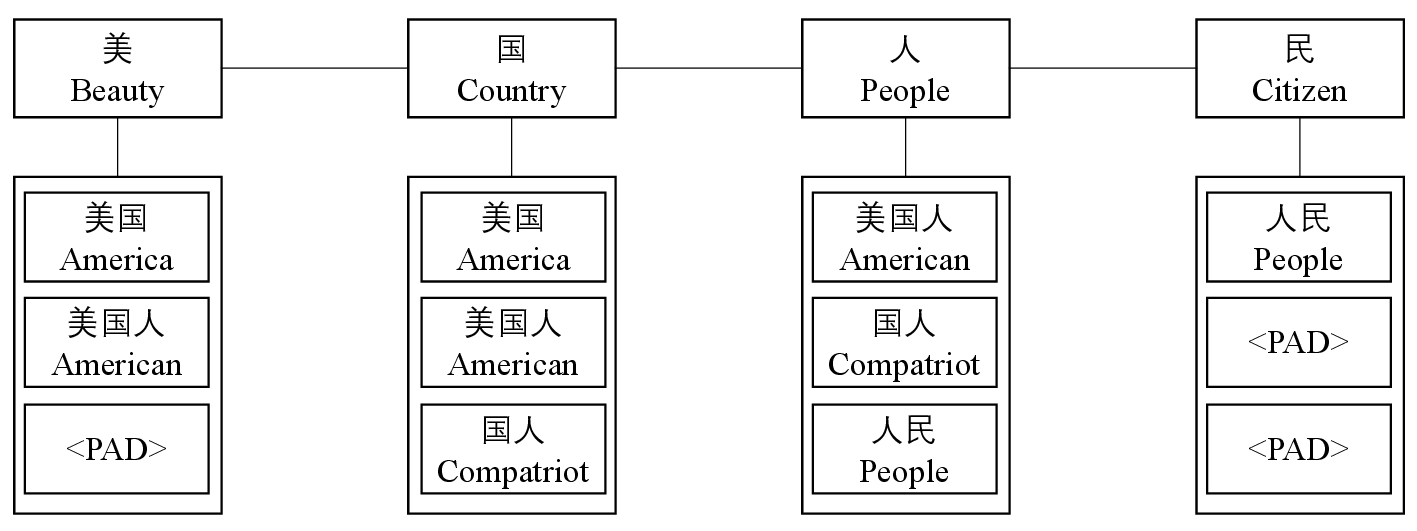}
\setlength{\abovecaptionskip}{1pt} 
\caption{Character-words pair sequence of a truncated Chinese sentence ``\pchn{美国人民} (American People)". There are four potential words, namely ``\pchn{美国} (America)'', ``\pchn{美国人} (American)", ``\pchn{国人} (Compatriot)", ``\pchn{人民} (People)". ``$<$PAD$>$" denotes padding value and each word is assigned to the characters it contains. }
\label{fig_char_words}
\vspace{-0.1in}
\end{figure}
To achieve it, we fine-tune the original parameters of BERT instead of fixing them, since directly injecting lexicon features into BERT will affect the performance due to the difference between that two information.




\section{Method}
The main architecture of the proposed Lexicon Enhanced BERT is shown in Figure \ref{fig_model}. Compared to BERT, LEBERT has two main differences. First, LEBERT takes both character and lexicon features as the input given that the Chinese sentence is converted to a character-words pair sequence. Second, a lexicon adapter is attached between Transformer layers, allowing lexicon knowledge integrated into BERT effectively.

In this section we describe: 1) Char-words Pair Sequence (Section \ref{char-word-section}), which incorporates words into a character sequence naturally; 2) Lexicon Adapter (Section \ref{la-section}), by injecting external lexicon features into BERT; 3) Lexicon Enhanced BERT (Section \ref{lebert-section}), by applying the Lexicon Adapter to BERT.


\subsection{Char-Words Pair Sequence}\label{char-word-section}
A Chinese sentence is usually represented as a character sequence, containing character-level features solely. To make use of lexicon information, we extend the character sequence to a character-words pair sequence. 

Given a Chinese Lexicon $\mathbf{D}$ and a Chinese sentence with $n$ characters $\mathbf{s_c} = \{c_1, c_2, ..., c_n\}$, we find out all the potential words inside the sentence by matching the character sequence with $\mathbf{D}$. Specifically, we first build a Trie based on the $\mathbf{D}$, then traverse all the character subsequences of the sentence and match them with the Trie to obtain all potential words. Taking the truncated sentence ``\pchn{美国人民} (American People)" for example, we can find out four different words, namely ``\pchn{美国} (America)", ``\pchn{美国人} (American)", ``\pchn{国人} (Compatriot)", ``\pchn{人民} (People)".  Subsequently, for each matched word, we assign it to the characters it contains. As shown in Figure \ref{fig_char_words},  the matched word ``\pchn{美国} (America)" is assigned to the character ``\pchn{美}" and ``\pchn{国}" since they form that word. Finally, we pair each character with assigned words and convert a Chinese sentence into a character-words pair sequence, i.e. $\mathbf{s_{cw}} = \{(c_1, ws_1), (c_2, ws_2), ..., (c_n, ws_n)\}$, where $c_i$ denotes the $i$-th character in the sentence and $ws_i$ denotes matched words assigned to $c_i$.

\subsection{Lexicon Adapter}\label{la-section}
\begin{figure}[t]
\centering\includegraphics[scale=0.4,trim=1 0 0 0]{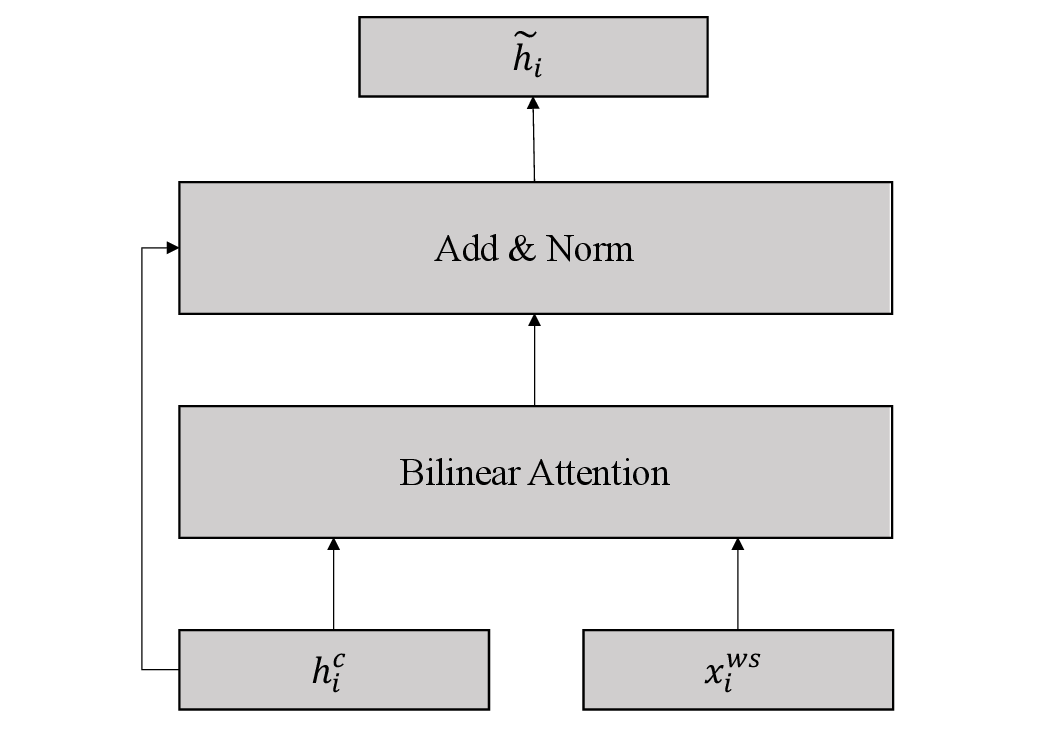}
\setlength{\abovecaptionskip}{1pt} 
\caption{Structure of Lexicon Adapter (LA). The adapter inputs a character vector and the paired word features. The bilinear attention over both character and words is used to weighted the lexicon feature into a vector, which is then added to the character-level vector and followed by a layer normalization.}
\label{fig_leixcon}
\vspace{-0.1in}
\end{figure}

Each position in the sentence consists of two types of information, namely character-level and word-level features. In line with the existing hybrid models, our goal is to combine the lexicon feature with BERT. Specifically, inspired by the recent works about BERT adapter ~\citep{houlsby2019parameter,wang2020kadapter}, we propose a novel Lexicon Adapter (LA) shown in Figure \ref{fig_leixcon}, which can directly inject lexicon information into BERT.

A Lexicon Adapter receives two inputs, a character and the paired words. For the $i$-th position in a char-words pair sequence, the input is denoted as $(h_i^c, x_i^{ws})$, where $h_i^c$ is a character vector, the output of a certain transformer layer in BERT, and $x_i^{ws} = \{x_{i1}^w, x_{i2}^w,..., x_{im}^w\}$ is a set of word embeddings. The $j$-th word in $x_i^{ws}$ is represented as following:
\begin{equation}
    x_{ij}^w = {\rm \mathbf{e}}^w(w_{ij}) 
\end{equation}
where ${\rm \mathbf{e}}^w$ is a pre-trained word embedding lookup table and $w_{ij}$ is the $j$-th word in $ws_i$.

To align those two different representations, we apply a non-linear transformation for the word vectors:
\begin{equation}
    v_{ij}^w = {\rm \mathbf{W}}_2({\rm tanh}({\rm \mathbf{W}}_{1} x_{ij}^{w} \;+ \; {\rm \mathbf{b}}_{1})) + {\rm \mathbf{b}}_2
\end{equation}
where ${\rm \mathbf{W}}_1$ is a $d_c$-by-$d_w$ matrix, ${\rm \mathbf{W}}_2$ is a $d_c$-by-{$d_c$} matrix, and ${\rm \mathbf{b}}_1$ and ${\rm \mathbf{b}}_2$ are scaler bias. $d_w$ and $d_c$ denote the dimension of word embedding and the hidden size of BERT respectively. 


As Figure \ref{fig_char_words} shows, each character is paired with multiple words. However, the contribution to each task varies from word to word. For example, as for Chinese POS tagging, words ``\pchn{美国} (America)" and ``\pchn{人民} (People)" are superior to ``\pchn{美国人} (American)" and ``\pchn{国人} (Compatriot)", since they are ground-truth segmentation of the sentence. To pick out the most relevant words from all matched words, we introduce a character-to-word attention mechanism.

Specifically,  we denote all $v_{ij}^w$ assigned to $i$-th character as $V_i = (v_{i1}^w, ...,v_{im}^w)$, which has the size $m$-by-$d_c$ and $m$ is the total number of the assigned word. The relevance of each word can be calculated as:
\begin{equation}
    \mathbf{a}_i = {\rm softmax}(h_i^c{\rm \mathbf{W}}_{attn}{V_i}^T)
\end{equation}
where ${\rm \mathbf{W}}_{attn}$ is the weight matrix of bilinear attention. Consequently, we can get the weighted sum of all words by:
\begin{equation}
    z_i^w = \sum_{j=1}^{m}a_{ij}v_{ij}^w
\end{equation}

Finally, the weighted lexicon information is injected into the character vector by:
\begin{equation}
\tilde{h}_i = h_i^c \; + \; z_i^w 
\end{equation}
It is followed by a dropout layer and layer normalization.

\subsection{Lexicon Enhanced BERT}\label{lebert-section}
Lexicon Enhanced BERT (LEBERT) is a combination of Lexicon Adapter (LA) and BERT, in which LA is applied to a certain layer of BERT shown in Figure \ref{fig_model}. Concretely, LA is attached between certain transformers within BERT, thereby injecting external lexicon knowledge into BERT.

Given a Chinese sentence with n characters $\mathbf{s}_c = \{c_1, c_2, ..., c_n\}$, we build the corresponding character-words pair sequence $\mathbf{s}_{cw} = \{(c_1, ws_1), (c_2, ws_2), ..., (c_n, ws_n)\}$ as described in Section \ref{char-word-section}. The characters $\{c_1, c_2, ..., c_n\}$ are first input into Input Embedder which outputs $E = \{e_1, e_2, ..., e_n\}$ by adding token, segment and position embedding. Then we input $E$ into Transformer encoders and each Transformer layer acts as following: 

\begin{equation}
\begin{aligned}
G & = {\rm LN}(H^{l-1} + {\rm MHAttn}(H^{l-1})) \\
H^{l} & = {\rm LN}(G + {\rm FFN}(G)) \\
\end{aligned}
\end{equation}
where $H^l = \{h_1^{l}, h_2^{l}, ..., h_n^{l}\}$ denotes the output of the $l$-th layer and $H^0 = E$; LN is layer normalization; MHAttn is the multi-head attention mechanism; FFN is a two-layer feed-forward network with ReLU as hidden activation function.

To inject the lexicon information between the $k$-th and $(k+1)$-th Transformer, we first get the output $H^k =  \{h_1^k, h_2^k, ..., h_n^k\}$ after $k$ successive Transformer layers. Then, each pair $(h_i^k, x_i^{ws})$ are passed through the \textbf{Lexicon Adapter} which transforms the $i_{th}$ pair into $\tilde{h}_i^k$:
\begin{equation}
    \tilde{h}_i^k = {\rm LA}(h_i^k, x_i^{ws})
\end{equation}

Since there are $L=12$ Transformer layers in the BERT, we input $\widetilde{H}^k = \{\tilde{h}_1^k, \tilde{h}_2^k, ..., \tilde{h}_n^k\}$ to the remaining $(L-k)$ Transformers. At the end, we get the output of $L$-th Transformer $H^L$ for the sequence labeling task.

\subsection{Training and Decoding}
Considering the dependency between successive labels, we use a CRF layer to make sequence labeling. Given the hidden outputs of the last layer $H^L = \{h_1^L, h_2^L, ..., h_n^L\}$, we first calculate scores $P$ as:
\begin{equation}
    O = {\rm \mathbf{W}}_o H^L + {\rm \mathbf{b}}_o
\end{equation}
For a label sequence $\mathbf{y}=\{y_1, y_2, ..., y_n\}$, we define its probability to be:
\begin{equation}
    p(\mathbf{y} | \mathbf{s}) = \frac{{\rm exp}(\sum_i(O_{i, y_i} + T_{y_{i-1}, y_i}))}{\sum_{\tilde{\mathbf{y}}}{\rm exp}(\sum_i(O_{i, \tilde{y}_i} + T_{\tilde{y}_{i-1}, \tilde{y}_i}))}
\end{equation}
where T is the transition score matrix and $\mathbf{\tilde{y}}$ denotes all possible tag sequences.

Given $N$ labelled data $\{\mathbf{s}_j, \mathbf{y}_j\}|_{j=1}^N$, we train the model by minimize the sentence-level negative log-likelihood loss as:
\begin{equation}
    \mathcal{L} = -\sum_{j}\;{\rm log}(p(\mathbf{y} | \mathbf{s}))
\end{equation}

While decoding, we find out the label sequence obtaining the highest score using the Viterbi algorithm.

\begin{table}[t]
\centering
\scalebox{0.8}{
\begin{tabular}{|l|l|l|l|l|l|}
\hline
\multicolumn{2}{|l|}{Dataset}                     & Type & Train   & Dev    & Test   \\ \hline
\multirow{8}{*}{NER} & \multirow{2}{*}{Weibo}     & Sent & 1.4k    & 0.27k  & 0.27k  \\ \cline{3-6} 
                     &                            & Char & 73.8k   & 14.5k  & 14.8k  \\ \cline{2-6} 
                     & \multirow{2}{*}{Ontonotes} & Sent & 15.7k   & 4.3k   & 4.3k   \\ \cline{3-6} 
                     &                            & Char & 491.9k  & 200.5k & 208.1k \\ \cline{2-6} 
                     & \multirow{2}{*}{MSRA}      & Sent & 46.4k   & -      & 4.4k   \\ \cline{3-6} 
                     &                            & Char & 2169.9k & -      & 172.6k \\ \cline{2-6} 
                     & \multirow{2}{*}{Resume}    & Sent & 3.8k    & 0.46k  & 0.48k  \\ \cline{3-6} 
                     &                            & Char & 124.1k  & 13.9k  & 15.1k  \\ \hline
\multirow{6}{*}{CWS} & \multirow{2}{*}{PKU}       & Sent & 19.1k   & -      & 1.9k   \\ \cline{3-6} 
                     &                            & Char & 1826k   & -      & 173k   \\ \cline{2-6} 
                     & \multirow{2}{*}{MSR}       & Sent & 86.9k   & -      & 4.0k   \\ \cline{3-6} 
                     &                            & Char & 4050k   & -      & 184k   \\ \cline{2-6} 
                     & \multirow{2}{*}{CTB6}      & Sent & 23k     & 2k     & 3k     \\ \cline{3-6} 
                     &                            & Char & 1056k   & 100k   & 134k   \\ \hline
\multirow{6}{*}{POS} & \multirow{2}{*}{CTB5}      & Sent & 18k     & 350    & 348    \\ \cline{3-6} 
                     &                            & Char & 805k    & 12k    & 14k    \\ \cline{2-6} 
                     & \multirow{2}{*}{CTB6}      & Sent & 23k     & 2k     & 3k     \\ \cline{3-6} 
                     &                            & Char & 1056k   & 100k   & 134k   \\ \cline{2-6} 
                     & \multirow{2}{*}{UD}        & Sent & 4k      & 500    & 500    \\ \cline{3-6} 
                     &                            & Char & 156k    & 20k    & 19k    \\ \hline
\end{tabular}
}
\caption{The statistics of the datasets.}
\label{table-statistics}
\end{table}

\section{Experiments}
We carry out an extensive set of experiments to investigate the effectiveness of LEBERT. In addition, we aim to empirically compare model-level and BERT-level fusion in the same setting. Standard F1-score (F1) is used as evaluation metrics.
\subsection{Datasets}
We evaluate our method on ten datasets of three different sequence labeling tasks, including Chinese NER, Chinese Word Segmentation, and Chinese POS tagging. The statistics of the datasets is shown in Table \ref{table-statistics}.

\noindent \textbf{Chinese NER}. We conduct experiments on four benchmark datasets, including Weibo NER ~\citep{peng-dredze-2015-named, peng-dredze-2016-improving}, OntoNotes ~\citep{weischedel2011ontonotes}, Resume NER ~\citep{zhang-yang-2018-chinese}, and MSRA ~\citep{levow2006third}. Weibo NER is a social media domain dataset, which is drawn from Sina Weibo; while OntoNotes and MSRA datasets are in the news domain. Resume NER dataset consists of resumes of senior executives, which is annotated by \citet{zhang-yang-2018-chinese}.

\noindent \textbf{Chinese Word Segmentation}. For Chinese word segmentation, we employ three benchmark datasets in our experiments, namely PKU, MSR, and CTB6, where the former two are from SIGHAN 2005 Bakeoff \cite{emerson-2005-second} and the last one is from \citet{xue2005penn}. For MSR and PKU, we follow their official training/test data split. For CTB6, we use the same split as that stated in \citet{yang-xue-2012-chinese, higashiyama-etal-2019-incorporating}.

\noindent \textbf{Chinese POS Tagging}. For POS-tagging, three Chinese benchmark datasets are used, including CTB5 and CTB6 from the Penn Chinese TreeBank \cite{xue2005penn} and the Chinese GSD Treebank of Universal Dependencies(UD) \cite{nivre-etal-2016-universal}. The CTB datasets are in simplified Chinese while the UD dataset is in traditional Chinese. Following \citet{shao-etal-2017-character}, we first convert the UD dataset into simplified Chinese before the POS-tagging experiments\footnote{The conversion tool we used is \href{https://github.com/BYVoid/OpenCC}{OpenCC}.}. Besides, UD has both universal and language-specific POS tags, we follow previous works ~\citep{shao-etal-2017-character,tian-etal-2020-joint-chinese}, referring to the corpus with two tagsets as UD1 and UD2, respectively. We use the official splits of train/dev/test in our experiments.

\subsection{Experimental Settings}
Our model is constructed based on ${\rm BERT}_{BASE}$ \cite{devlin-etal-2019-bert}, with 12 layers of transformer, and is initialized using the Chinese-BERT checkpoint from huggingface\footnote{https://github.com/huggingface/transformers}. We use the 200-dimension pre-trained word embedding from \citet{song-etal-2018-directional}, which is trained on texts of news and webpages using a directional skip-gram model. The lexicon $\mathbf{D}$ used in this paper is the vocab of the pre-trained word embedding. We apply the Lexicon Adapter between the $1$-st and $2$-nd Transformer in BERT and fine-tune both BERT and pre-trained word embedding during training.

\noindent{\textbf{Hyperparameters}}. We use the Adam optimizer with an initial learning rate of 1e-5 for original parameters of BERT, and 1e-4 for other parameters introduced by LEBERT, and a maximum epoch number of 20 for training on all datasets. The max length of the sequence is set to 256, and the training batch size is 20 for MSRA NER and 4 for other datasets.

\noindent{\textbf{Baselines}}. To evaluate the effectiveness of the proposed LEBERT, we compare it with the following approaches in the experiments.
\begin{itemize}[leftmargin=*]
    \item \textbf{BERT}. Directly fine-tuning a pre-trained Chinese BERT on Chinese sequence labeling tasks.
    \item \textbf{BERT+Word}. A strong model-level fusion baseline method, which inputs the concatenation of BERT vector and bilinear attention weighted word vector, and uses LSTM\footnote{We also evaluated with other fusion layers, such as Transformer, but we found LSTM is consistently better.} and CRF as fusion layer and inference layer respectively.
    \item \textbf{ERNIE} \cite{sun2019ernie}. An extension of BERT using a entity-level mask to guide pre-training.
    \item \textbf{ZEN}. \citet{diao-etal-2020-zen} explicitly integrate N-gram information into BERT through an extra multi-layers of N-gram Transformer encoder and pre-training.
\end{itemize}

Further, we also compare with the state-of-the-art models of each task. 
\begin{table}[]
\centering
\scalebox{0.70}{
\begin{tabular}{lcccc}
\hline
Model       & Weibo & Ontonotes & MSRA  & Resume \\ \hline
\citet{zhang-yang-2018-chinese}\textsuperscript{*}     & 63.34 & 75.49     & 92.84 & 94.51  \\
\citet{zhu-wang-2019-ner}     & 59.31 & 73.64     & 92.97 & 94.94  \\
\citet{liu-etal-2019-encoding}\textsuperscript{*}      & 65.30 & 75.79     & 93.50 & 94.49  \\
\citet{ding-etal-2019-neural}     & 59.50 & 75.20     & 94.40 & -  \\
\citet{ma-etal-2020-simplify}\textsuperscript{* ${\dagger}$}      & 69.11 & 81.34     & 95.35 & 95.54  \\
\citet{li-etal-2020-flat}\textsuperscript{* ${\dagger}$}        & 68.07 & 80.56     & 95.46 & 95.78  \\ \hline \hline
BERT        & 67.27 & 79.93     & 94.71 & 95.33  \\
BERT+Word & 68.32 & 81.03     & 95.32 & 95.46  \\
ERINE       & 67.96     & 77.65         & 95.08     & 94.82      \\
ZEN         &66.71     & 79.03         & 95.20  & 95.40      \\ \hline \hline
LEBERT      & \textbf{70.75} & \textbf{82.08}     & \textbf{95.70} & \textbf{96.08}  \\ \hline 
\end{tabular}
}
\caption{Results on Chinese NER.}
\label{table-ner}
\end{table}

\subsection{Overall Results}
\noindent{\textbf{Chinese NER}}. Table \ref{table-ner} shows the experimental results on Chinese NER datasets\footnote{For a fair comparison, in Table \ref{table-ner}, we use * denotes training the model with the same pre-trained word embedding as ours; ${\dagger}$ means the model is also initialized using the Chinese BERT checkpoint from huggingface and evaluated using the \href{https://github.com/chakki-works/seqeval}{seqeval} tool. }. The first four rows~\citep{zhang-yang-2018-chinese,zhu-wang-2019-ner,liu-etal-2019-encoding,ding-etal-2019-neural} in the first block show the performance of lexicon enhanced character-based Chinese NER models, and the last two rows~\citep{ma-etal-2020-simplify,li-etal-2020-flat} in the same block are the state-of-the-art models using shallow fusion layer to integrate lexicon information and BERT. The hybrid models, including existing state-of-the-art models, BERT + Word, and the proposed LEBERT, achieve better performance than both lexicon enhanced models and BERT baseline. This demonstrates the effectiveness of combining BERT and lexicon features for Chinese NER. Compared with model-level fusion models (\cite{ma-etal-2020-simplify,li-etal-2020-flat}, and BERT+Word), our BERT-level fusion model, LEBERT, improves in F1 score on all four datasets across different domains, which shows that our approach is more efficient in integrating word and BERT. The results also indicate that our adapter-based method, LEBERT, with an extra pre-trained word embedding solely, outperforms those two lexicon-guided pre-training models (ERNIE and ZEN). This is likely because implicit integration of lexicon in ERNIE and restricted pre-defined n-gram vocabulary size in ZEN limited the effect.

\begin{table}[]
\centering
\scalebox{0.8}{
\begin{tabular}{lccc}
\hline
Model                & PKU   & MSR   & CTB6  \\ \hline
\citet{yang-etal-2017-neural}             & 95.00  & 96.80  & 95.40  \\
\citet{ma-etal-2018-state}               & 96.10  & 97.40  & 96.70  \\
\citet{yang-etal-2019-subword}             & 95.80  & 97.80  & 96.10  \\
\citet{qiu-etal-2020-concise}              & 96.41 & 98.05 & 96.99     \\
\citet{tian-etal-2020-improving}(with BERT) & 96.51 & 98.28 & 97.16 \\
\citet{tian-etal-2020-improving}(with ZEN)  & 96.53 & 98.40 & 97.25 \\ \hline \hline
BERT                 & 96.25     & 97.94     & 96.98     \\
BERT+Word          & 96.55     & 98.41     & 97.25     \\
ERINE                & 96.33     & 98.17     & 97.02     \\
ZEN                  & 96.36     & 98.36     & 97.13     \\ \hline \hline
LEBERT               & \textbf{96.91} & \textbf{98.69} & \textbf{97.52} \\ \hline
\end{tabular}}
\caption{Results on Chinese Word Segmentation.}
\label{table-cws}
\end{table}

\noindent \textbf{Chinese Word Segmentation}. We report the F1 score of our model and the baseline methods on Chinese Word Segmentation in Table \ref{table-cws}. \citet{yang-etal-2019-subword} applied a lattice LSTM to integrate word feature to character-based CWS model. \citet{qiu-etal-2020-concise} investigated the benefit of multiple heterogeneous segmentation criteria for single criterion Chinese word segmentation. \citet{tian-etal-2020-improving} designed a wordhood memory network to incorporate wordhood information into a pre-trained-based CWS model and showed good performance. Compared with those approaches, the models (BERT+Word and LEBERT) that combine lexicon features and BERT perform better. Moreover, our proposed LEBERT outperforms both model-level fusion baseline (BERT+Word) and lexicon-guided pre-training models (ERNIE and ZEN), achieving the best results.

\begin{table}[]
\centering
\scalebox{0.75}{
\begin{tabular}{lcccc}
\hline
Model          & CTB5  & CTB6  & UD1   & UD2   \\ \hline
\citet{shao-etal-2017-character}       & 94.38 & -     & 89.75 & 89.42 \\
\citet{8351918}      & 94.95 & 92.51 & -     & -     \\
\citet{tian-etal-2020-joint-chinese}(BERT) & 96.77 & 94.82 & 95.51 & 95.46 \\
\citet{tian-etal-2020-joint-chinese}(ZEN)  & 96.86 & 94.87 & 95.52 & 95.49 \\
\citet{tian-etal-2020-joint}(BERT) & 96.60 & 94.74 & 95.50 & 95.38 \\
\citet{tian-etal-2020-joint}(ZEN) & 96.82 & 94.82 & 95.59 & 95.41 \\ \hline \hline
BERT           & 96.25     & 94.64     & 94.83     & 94.73     \\
BERT+Word      & 96.77     & 94.75     & 95.39     & 95.41     \\
ERINE          & 96.51     & 94.76     & 95.10     & 95.14     \\
ZEN            & 96.60     & 94.70     & 95.15     & 95.05     \\ \hline \hline
LEBERT         & \textbf{97.14} & \textbf{95.18} & \textbf{96.06} & \textbf{95.74} \\ \hline
\end{tabular}}
\caption{Results on Chinese POS Tagging.}
\label{table-pos}
\end{table}

\begin{table}[]
\centering
\scalebox{0.9}{
\begin{tabular}{llcc}
\hline
                     &           & BERT    & STOA (with BERT) \\ \hline
\multirow{4}{*}{NER} & Weibo     & 10.63\% & 5.31\%            \\ 
                     & Ontonote4 & 10.71\%  & 3.97\%           \\  
                     & MSRA      & 18.71\%  & 5.28\%            \\ 
                     & Resume    & 16.06\%  & 7.11\%            \\ \hline
\multirow{3}{*}{CWS} & PKU       & 17.60\%  & 11.46\%           \\  
                     & MSR       & 36.41\%  & 23.84\%           \\  
                     & CTB6      & 17.88\%  & 12.68\%           \\ \hline
\multirow{4}{*}{POS} & CTB5      & 23.73\%  & 11.46\%           \\ 
                     & CTB6      & 10.07\%  & 6.95\%            \\  
                     & UD1       & 23.79\%  & 12.25\%           \\ 
                     & UD2       & 19.17\%  & 6.17\%            \\ \hline
\end{tabular}}
\caption{The relative error reductions over different base models.}
\label{table-error}
\end{table}

\noindent \textbf{Chinese POS Tagging}. We report the F1 score on four benchmarks of Chinese POS tagging in Table \ref{table-pos}. The state-of-the-art model \cite{tian-etal-2020-joint-chinese} jointly trains Chinese Word Segmentation and Chinese POS tagging using a two-way attention to incorporate auto-analyzed knowledge, such as POS labels, syntactic constituents, and dependency relations. Similar to BERT+Word baseline, \citet{tian-etal-2020-joint} integrated character-Ngram features with BERT at model-level using a multi-channel attention. As shown in Table \ref{table-pos}, hybrid models (\cite{tian-etal-2020-joint}, BERT+Word, LEBERT) that combine words information and BERT outperform BERT baseline, indicating that lexicon features can further improve the performance of BERT. LEBERT achieves the best results among these approaches, which demonstrates the effectiveness of BERT-level fusion. Consistent with results on Chinese NER and CWS, our BERT adapter-based approach is superior to lexicon-guided pre-training methods (ERNIE and ZEN). 

Our proposed model has achieved state-of-the-art results across all datasets. To better show the strength of our method, we also summarize the relative error reduction over BERT baseline and BERT-based state-of-the-art models in Table \ref{table-error}. The results show that the relative error reductions are significant compared with baseline models.

\subsection{Model-level Fusion vs. BERT-level Fusion}
Compared with model-level fusion models, LEBERT directly integrates lexicon features into BERT. We evaluate those two types of models in terms of Span F1, Type Acc, and Sentence Length, choosing the BERT+Word as the model-level fusion baseline due to its good performance across all the datasets. We also compare with a BERT baseline since both LEBERT and BERT+Word are improved based on it. 

\noindent \textbf{Span F1} \& \textbf{Type Acc}. Span F1 means the correctness of the span for an Entity in NER or a word in POS-tagging, while Type Acc denotes the proportion of full-correct predictions to span-correct predictions. Table \ref{table-span-type} shows the results of three models on the Ontonotes and UD1 datasets. We can find that both BERT+Word and LEBERT perform better than BERT in terms of Span F1 and Type Acc on the two datasets. The results indicate that lexicon information contributes to span boundary detection and span classification. Specifically, the improvement of Span F1 is larger than Type Acc on Ontonotes, but smaller on UD1. Compared with BERT+Word, LEBERT achieves more improvement, demonstrating the effectiveness of lexicon feature enhanced via BERT-level fusion.

\begin{table}[t]
\centering
\scalebox{0.82}{
\begin{tabular}{lcccc}
\hline
          & \multicolumn{2}{l}{\hspace{2em}Span F1} & \multicolumn{2}{l}{\hspace{2em}Type Acc} \\ \hline
          & Ontonotes        & UD1          & Ontonotes         & UD1          \\
BERT      & 82.68        & 97.99        & 97.16         & 96.99        \\
BERT+Word & 83.38        & 98.09        & 97.24         & 97.51        \\
LEBERT    & 84.16        & 98.47        & 97.84         & 97.72        \\ \hline  
\end{tabular}
}
\caption{Span F1 and Type Acc of different models.}
\label{table-span-type}
\end{table}

\begin{figure}[t]
\centering
\includegraphics[scale=0.41,trim=1 1 1 1]{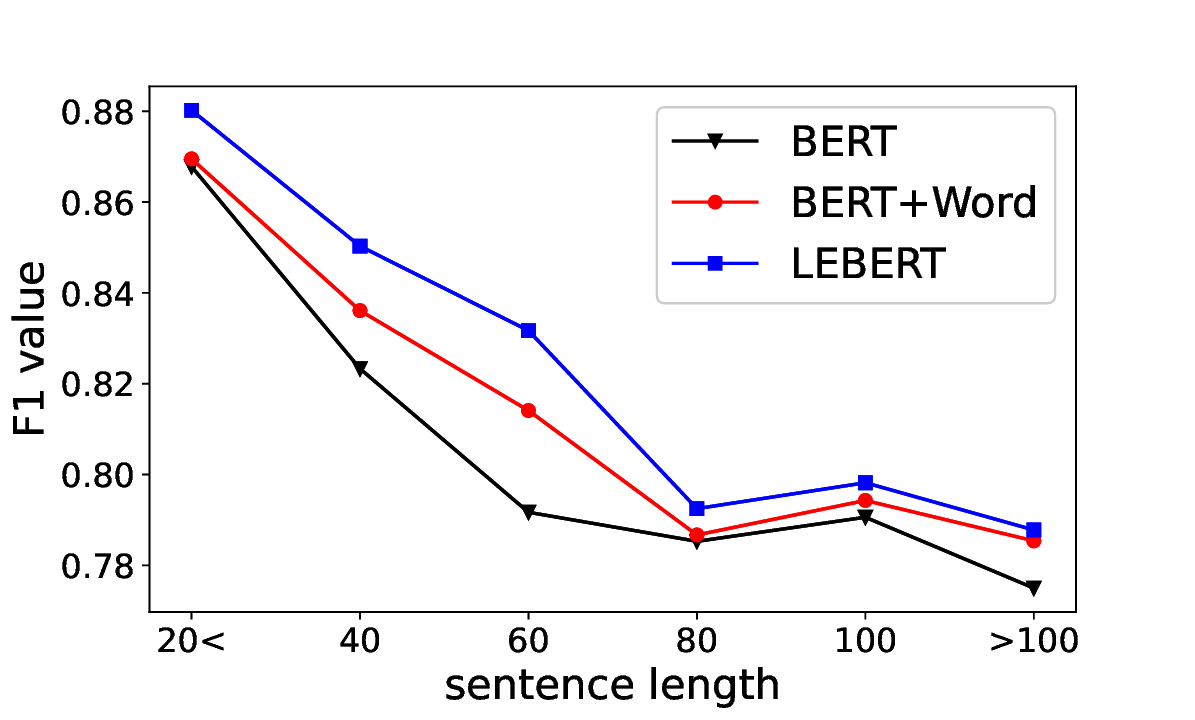}
\setlength{\abovecaptionskip}{1pt} 
\caption{F1-value against the sentence length.}
\label{fig_sent-len}
\vspace{-0.1in}
\end{figure}

\noindent \textbf{Sentence Length}. Figure \ref{fig_sent-len} shows the F1-value trend of the baselines and LEBERT on Ontonotes dataset. All the models show a similar performance-length curve, decreasing as the sentence length increase. We speculate that long sentences are more challenging due to complicated semantics. Even lexicon enhanced models may fail to choose the correct words because of the increased number of matched words as the sentence become longer. The F1-score of BERT is relatively low, while BERT+Word achieves better performance due to the usage of lexicon information. Compared with BERT+Word, LEBERT performs better and shows more robustness when sentence length increases, demonstrating the more effective use of lexicon information.

\noindent \textbf{Case Study}. Table \ref{table-case} shows examples of Chinese NER and Chinese POS tagging results on Ontonotes and UD1 datasets respectively. In the first example, BERT can not determine the entity boundary, but BERT+Word and LEBERT can segment it correctly. However, the BERT+Word model fails to predict the type of the entity ``\pchn{呼伦贝尔盟} (Hulunbuir League)" while LEBERT makes the correct prediction. This is likely because fusion at the lower layer contributes to capturing more complex semantics provided by BERT and lexicon. In the second example, the three models can find the correct span boundary, but both BERT and BERT+Word make incorrect predictions of the span type. Although BERT+Word can use the word information, it is disturbed by the irrelevant word ``\pchn{七八} (Seven and Eight)" predicting it as NUM. In contrast, LEBERT can not only integrate lexicon features but also choose the correct word for prediction.

\begin{table}[t]
\centering
\scalebox{0.85}{
\begin{tabular}{|l|c|c|c|c|c|}
\hline
\multirow{2}{*}{Layer} & \multicolumn{5}{l|}{\hspace{7.5em}\textbf{\textit{one}}}                                \\ \cline{2-6} 
                       & 1      & 3      & 6        & 9            & 12          \\ \hline
F1                     & 82.08  & 81.43  & 81.24    & 81.10        & 80.64       \\ \hline
\multirow{2}{*}{Layer} & \multicolumn{3}{l|}{\hspace{3.6em}\textbf{\textit{multi}}} & \multicolumn{2}{l|}{\hspace{2.2em}\textbf{\textit{all}}}   \\ \cline{2-6} 
                       & 1,3    & 1,3,6  & 1,3,6,9  & \multicolumn{2}{c|}{all}   \\ \hline
F1                     & 81.54  & 81.28  & 81.23    & \multicolumn{2}{c|}{78.54} \\ \hline
\end{tabular}}
\caption{Results of variations of LEBERT with Lexicon Adapter applied at different layers of BERT model. \textbf{\textit{one}}, \textbf{\textit{multi}}, \textbf{\textit{all}} mean applying LA after one layer, multiply layers, all layers of Transformer in BERT.}
\vspace{-6pt}
\label{table-layers}
\end{table}

\begin{table*}[t]
\centering
\scalebox{0.72}{
\begin{tabular}{|l|cccccccc|}
\hline
\multicolumn{9}{|l|}{\#1 Example of Chinese NER}                                                                                                                                                                                                                                         \\ \hline
\begin{tabular}[c]{@{}l@{}}Sentence (truncated)\end{tabular} & \multicolumn{8}{l|}{\begin{tabular}[c]{@{}l@{}}\pchn{内蒙古呼伦贝尔盟}  (Hulunbuir League, Inner Mongolia)\end{tabular}}                                                                                                              \\ \hline
Matched Words                                                   & \multicolumn{8}{l|}{\begin{tabular}[c]{@{}l@{}}\pchn{内蒙}, \pchn{内蒙古}, \pchn{内蒙古呼伦贝尔}, \pchn{蒙古}, \pchn{呼伦}, \pchn{呼伦贝尔}, \pchn{呼伦贝尔盟}, \pchn{贝尔}\\ Inner Mongolia, Inner Mongolia, Inner Mongolia Hulunbuir, Mongolia, Hulun, \\ Hulunbuir, Hulunbuir League, Buir\end{tabular}} \\ \hline
Characters                                                     & \pchn{内}                         & \pchn{蒙}                         & \pchn{古}                         & \pchn{呼}                         & \pchn{伦}                        & \pchn{贝}                       & \pchn{尔}                       & \pchn{盟}                      \\ \cline{1-1} \hline
Gold Labels                                                     & B-GPE                     & I-GPE                     & E-GPE                     & B-GPE                     & I-GPE                    & I-GPE                   & I-GPE                   & E-GPE                  \\ \cline{1-1} \hline
BERT                                                           & B-GPE                     & I-GPE                     & I-GPE                     & I-GPE                     & I-GPE                    & I-GPE                   & I-GPE                   & E-GPE                  \\ \cline{1-1} \hline
BERT+Word                                                      & B-GPE                     & I-GPE                     & E-GPE                     & B-ORG                     & I-ORG                    & I-ORG                   & I-ORG                   & E-ORG                  \\ \cline{1-1} \hline
LEBERT                                                      & B-GPE                     & I-GPE                     & E-GPE                     & B-GPE                     & I-GPE                    & I-GPE                   & I-GPE                   & E-GPE                  \\ \hline \hline
\multicolumn{9}{|l|}{\#2 Example of Chinese POS Tagging}                                                                                                                                                                                                                                         \\ \hline
\begin{tabular}[c]{@{}l@{}}Sentence (truncated)\end{tabular} & \multicolumn{8}{l|}{\begin{tabular}[c]{@{}l@{}}\pchn{乱七八糟的关系} (Messy Relationship)\end{tabular}}                                                                                                           \\ \hline
Matched Words                                                  & \multicolumn{8}{l|}{\begin{tabular}[c]{@{}l@{}}\pchn{乱七八糟}, \pchn{七八}, \pchn{八糟}, \pchn{关系} \\ Mess, Seven and Eight, Bad News, Relationship\end{tabular}}                                                  \\ \hline
Characters                                                     & \pchn{乱}                         & \pchn{七}                         & \pchn{八}                         & \pchn{糟}                         & \pchn{的}                        & \pchn{关}                       & \pchn{系}                       &                        \\ \cline{1-1} \hline
Gold Labels                                                    & B-ADJ                    & I-ADJ                    & I-ADJ                    & E-ADJ                    & S-PART                   & B-NOUN                  & E-NOUN                  &                        \\ \cline{1-1} \hline
BERT                                                           & B-ADJ                   & I-NUM                   & I-NUM                    & E-ADJ                    & S-PART                   & B-NOUN                  & E-NOUN                  &                        \\ \cline{1-1} \hline
BERT+Word                                                      & B-ADJ                   & I-NUM                   & I-NUM                   & E-ADJ                   & S-PART                   & B-NOUN                  & E-NOUN                  &                        \\ \cline{1-1} \hline
LEBERT                                                         & B-ADJ                    & I-ADJ                    & I-ADJ                    & E-ADJ                    & S-PART                   & B-NOUN                  & E-NOUN                  &                        \\ \cline{1-1} \hline
\end{tabular}}
\caption{Examples of tagging result.}
\label{table-case}
\end{table*}

\subsection{Discussion}
\noindent \textbf{Adaptation at Different Layers}. We explore the effect of applying the Lexicon Adapter (LA) between different Transformer layers of BERT on Ontonotes dataset. Different settings are evaluated, including applying LA after \textit{one}, \textit{multiple}, and \textit{all} layers of Transformer. As for \textit{one} layer, we applied LA after $k \in \{1, 3, 6, 9, 12\}$ layer; and $\{1, 3\}$, $\{1, 3, 6\}$, $\{1, 3, 6, 9\}$ layers for \textit{multiple} layers. \textit{All} layers represents LA used after every Transformer layer in BERT. The results show in Table \ref{table-layers}. The shallow layer achieves better performance, which can be due to the fact that the shallow layer promotes more layered interaction between lexicon features and BERT. Applying LA at multi-layers of BERT hurts the performance and one possible reason is that integration at multi-layers causes over-fitting.

\noindent \textbf{Tuning BERT or Not}. Intuitively, integrating lexicon into BERT without fine-tuning can be faster \cite{houlsby2019parameter} but with lower performance due to the different characteristics of lexicon feature and BERT (discrete representation vs. neural representation). To evaluate its impact, we conduct experiments with and without fine-tuning BERT parameters on Ontonotes and UD1 datasets. From the results, we find that without fine-tuning the BERT, the F1-score shows a decline of 7.03 points (82.08 $\to$ 75.05) on Ontonotes and 3.75 points (96.06 $\to$ 92.31) on UD1, illustrating the importance of fine-tuning BERT for our lexicon integration.

\section{Conclusion}
In this paper, we proposed a novel method to integrate lexicon features and BERT for Chinese sequence labeling, which directly injects lexicon information between Transformer layers in BERT using a Lexicon Adapter. Compared with model-level fusion methods, LEBERT allows in-depth fusion of lexicon features and BERT representation at BERT-level. Extensive experiments show that the proposed LEBERT achieves state-of-the-art performance on ten datasets of three Chinese sequence labeling tasks.

\section*{Acknowledgments}
We would like to thank the anonymous reviewers for their valuable comments and suggestions. Moreover, We sincerely thank Dr. Zhiyang Teng for his constructive collaboration during the development of this paper, and Dr. Haixia Chai, Dr. Jie Yang, and my colleague Junfeng Tian for their help in polishing our paper. 


\bibliography{main.bbl}


\end{document}